\documentclass[conference]{IEEEtran}
\IEEEoverridecommandlockouts
\usepackage{cite}
\usepackage{amsmath,amssymb,amsfonts}
\usepackage{algorithmic}
\usepackage{graphicx}
\usepackage{textcomp}
\usepackage{xcolor}
\usepackage{multirow} 
\usepackage{booktabs}
\def\BibTeX{{\rm B\kern-.05em{\sc i\kern-.025em b}\kern-.08em
    T\kern-.1667em\lower.7ex\hbox{E}\kern-.125emX}}
\begin{document}

\title{Customizing Speech Recognition Model with Large Language Model Feedback
}

\author{
\IEEEauthorblockN{Shaoshi Ling}
\IEEEauthorblockA{\textit{Microsoft Core AI} \\
Redmond, WA, USA \\
shaoshiling@microsoft.com}
\and
\IEEEauthorblockN{Guoli Ye}
\IEEEauthorblockA{\textit{Microsoft Core AI} \\
Redmond, WA, USA \\
guoye@microsoft.com}
}

\maketitle

\begin{abstract}
Automatic speech recognition (ASR) systems have achieved strong performance on general transcription tasks. However, they continue to struggle with recognizing rare named entities and adapting to domain mismatches. In contrast, large language models (LLMs), trained on massive internet-scale datasets, are often more effective across a wide range of domains. In this work, we propose a reinforcement learning based approach for unsupervised domain adaptation, leveraging unlabeled data to enhance transcription quality—particularly the named entities affected by domain mismatch—through feedback from a LLM. Given contextual information, our framework employs a LLM as the reward model to score the hypotheses from the ASR model. These scores serve as reward signals to fine-tune the ASR model via reinforcement learning. Our method achieves a 21\% improvement on entity word error rate over conventional self-training methods.
\end{abstract}


\section{Introduction}
Automatic speech recognition (ASR) systems have made significant progress in recent years, driven by advances in neural network architectures and the availability of increasingly large amounts of labeled data \cite{li2022recent}. However, the performance of ASR systems often degrade significantly when the target domain or test conditions differ from those seen during training. And this domain mismatch between training and testing conditions occurs commonly in the real world application and poses a major challenge. One straightforward approach is to collect labeled data from the target domain and use it for adapting the pre-trained source model. However, manually annotating large volumes of data for each new domain is both costly and time-consuming. This highlights the need for the methods that can leverage unlabeled data to enable effective source-to-target domain transfer \cite{bell2020adaptation}.
 
The existing efforts to address this problem primarily focus on leveraging unlabeled target domain data to enhance the ASR performance. This approach is generally known as unsupervised domain adaptation and has been widely explored in speech processing communities. Among various techniques, the most widely used one is known as self-training\cite{raina2007self}, which typically involves two stages. First, a pre-trained model generates pseudo-labels for the target-domain data. Then, these pseudo-labeled samples, often accompanied by confidence scores, are used to adapt the model. The confidence scores are typically estimated using the pseudo posterior probabilities produced by the softmax function in ASR models\cite{li2021confidence}. However, these scores can be unreliable, especially in previously unseen domains.

In contrast, text large language models (LLMs) such as GPT4\cite{achiam2023gpt}, Deepseek\cite{guo2025deepseek}, Phi\cite{abdin2024phi} trained on internet-scale corpora, demonstrate remarkable generalization capabilities across a wide range of domains. These models have demonstrated strong performance on tasks involving commonsense reasoning, in-context learning, and even zero-shot adaptation. Such capabilities suggest that LLMs could serve as valuable resources for enhancing the performance of ASR models, particularly in low-resource or domain-mismatched settings. Recent studies have explored integrating LLMs with ASR systems through approaches such as re-ranking hypotheses\cite{hu2023massively, li2023prompting, chen2023large, ogawa2024applying} or error correction during post-processing\cite{ma2023can, gu2024denoising, ma2025asr}. 


Inspired by reinforcement learning (RL) from AI feedback \cite{bai2022constitutional, guo2024direct}, we propose an unsupervised domain adaptation of ASR systems using online LLMs feedback. Rather than relying on a separately trained reward model tailored for the ASR task, our method bypasses this requirement by directly extracting evaluation scores from an LLM. Specifically, given hypothesis of spoken utterance in the target domain along with contextual information (e.g., domain metadata, description of the targeted named entities or preceding dialogue), we use an LLM to evaluate the quality of those hypotheses by computing the sum of log-probability scores in the LLMs. These scores are then used to generate reward signals, which guide the fine-tuning of the ASR model using state-of-the-art RL algorithms like DPO\cite{rafailov2023direct, guo2024direct} and GRPO\cite{shao2024deepseekmath}. Unlike traditional self-training techniques, our method does not require confidence score. Instead, it leverages the LLMs' implicit understanding of language and context to produce reliable rewards, thereby promoting more accurate and contextually appropriate transcriptions in target domains.

Our experiments demonstrate that our RL framework can substantially improve transcription quality in domain-mismatched scenarios, particularly for named entities and context-sensitive phrases. Compared to standard self-training domain adaptation methods, our approach achieves up to 21\% relative improvement on entity word error rate (EWER), showcasing the effectiveness of LLMs feedback in guiding ASR model adaptation. Our contributions can be summarized as follows:
\begin{itemize}
\item We proposed novel reward function for RL in ASR tasks. Unlike prior approaches that rely on a separately trained reward model or human feedback, our method introduces a straightforward reward computation. Furthermore, our reward signal is dense and avoids common issues such as reward hacking, making it both effective and robust.
\item We applied the state-of-the-art RL algorithms to achieve contextual customization for ASR in the unsupervised domain adaptation setting.
\item By leveraging text LLMs, our approach yields up to a 21\% relative improvement on EWER compared with the standard self-training approach.
\end{itemize}

\begin{figure*}[pht]
\centering
{\includegraphics[width=0.8\textwidth,height=0.30\textwidth]{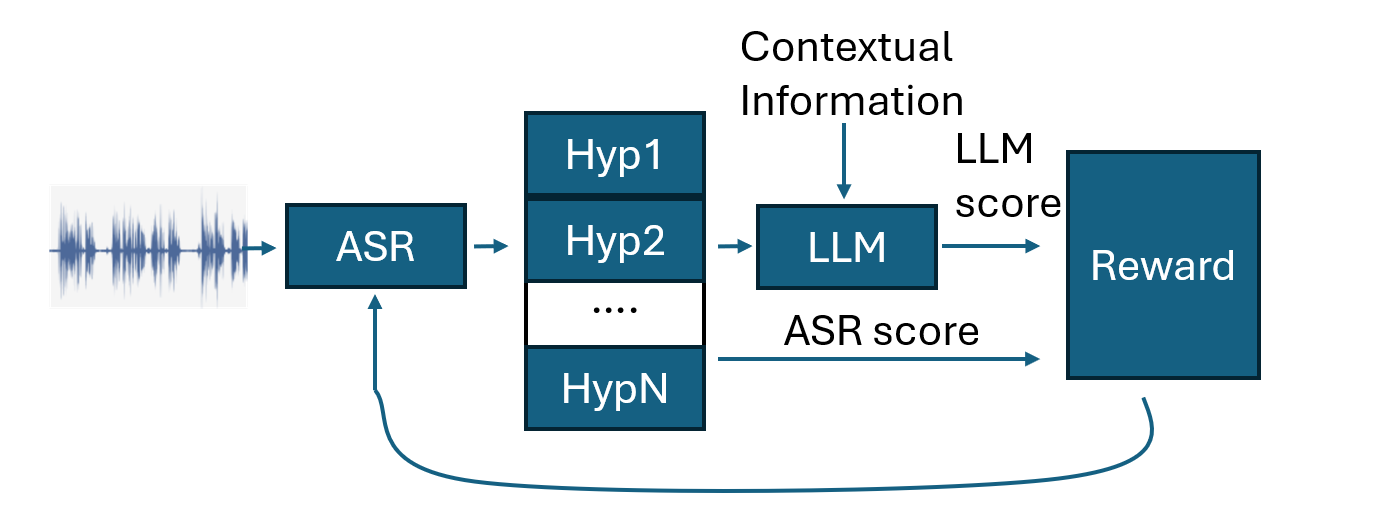}} %
\caption{The Overview}
\label{fig:overview}
\end{figure*}

\section{Related Work}

\subsection{Unsupervised Domain Adaptation in ASR}

Since obtaining ground-truth speech transcriptions in the target domain is often prohibitively expensive, many existing approaches leverage out-of-domain data to bootstrap improved models for the target domain\cite{zhu2023boosting, sun2017unsupervised}. Other methods includes simulating the target domain speech \cite{hosseini2018multi}, domain-invariant representation learning \cite{narayanan2018toward}, domain adversarial learning \cite{sun2017unsupervised} and teacher-student learning for efficient adaptation \cite{li2017large, meng2019conditional}. More recently, self-supervised pre-trained models have also been used for pseudo-labelling to achieve unsupervised adaptation \cite{khurana2021unsupervised, hwang2022large}. 
To address the inherent uncertainty in ASR decoding, a complementary line of research focuses on confidence estimation and data selection, which have been studied for decades. Recent efforts focus on performing pseudo-labeling denoising technique, such as uncertainty driven self-training \cite{khurana2021unsupervised}, mask prediction \cite{ling2022improving}, and leveraging internal model signals—e.g., analyzing the self-attention matrix to evaluate pseudo-label quality \cite{hu2024self}, or partitioning attention weights \cite{lee2023partitioning} to better fit the model on pseudo-label data.

\subsection{Reinforcement Learning in ASR}
In the field of ASR, several reinforcement learning approaches have been proposed to bridging the gap between the training and testing phases. Self-critical sequence training \cite{chen2022self} utilizes a word error rate based reward to reduce the reliance on teacher forcing during training. Others\cite{tjandra2018sequence} employed policy gradient method to train the S2S ASR model, treating sequence generation as a sequential decision-making process under the RL framework. And the work\cite{kala2018reinforcement} introduced a framework that incorporates user feedback through hypothesis selection, allowing the model to refine its predictions based on human preference.

While prior works have primarily relied on either human feedback or WER for designing reward signals, our approach explores the use of feedback from LLMs. To the best of our knowledge, this is the first work to apply RL in the contextual customization task for ASR. This opens up new possibilities for adapting ASR systems to diverse domains by utilizing powerful LLMs not merely in post-processing stages such as re-ranking\cite{hu2023massively, li2023prompting, chen2023large, ogawa2024applying} or error correction\cite{ma2023can, gu2024denoising, ma2025asr}, but as an integral part of the training process.

\section{Method}\label{AA}

\subsection{Adaptation Framework}
Given an utterance audio $X$, an ASR model can be viewed as a policy $\pi$ that maps the audio input into the corresponding transcription $Y$, represented as $\pi(Y|X)$. We define a reward function $r(Y|X)$ that assigns a feedback score to the transcription hypothesis. For each audio X, we can generate n candidate hypotheses: $a_1, a_2, \ldots, a_n$. 


As shown in Figure \ref{fig:overview}, our framework followed the general RL pipeline \cite{xiong2025minimalist} and consists of the following three steps:
\begin{itemize}
\item \textbf{Data collection.} For a batch of audio inputs ${X_1,\cdot\cdot\cdot,X_M}$, we sample n candidate responses for each $X_i$ from a pretrained ASR model (the reference model) resulting in hypothesis sets ${a_{i,1}, \cdot\cdot\cdot ,a_{i,n}}$.
\item \textbf{Generate Rewards.} For each audio $X_i$ and its contextual information given as a prompt $CP$, we compute the reward values of each hypothesis ${r_{i,1},\cdot\cdot\cdot ,r_{i,n}}$ using the reward model, which in our case is defined as the sum of log probabilities from the LLM conditioned on $CP$. The rewards are combined with scores from ASR model to compute advantages. The samples and their associated advantages are aggregated into the training dataset $\mathcal{D}$.
\item \textbf{Model Fine-tuning.} The policy model $\pi$ is then fine-tuned using RL algorithms over the training dataset $\mathcal{D}$.
\end{itemize}

\subsection{Reward Function}
In domain adaptation scenarios, contextual information such as domain metadata, preceding utterances, or descriptions of target named entities is often available. We incorporate this information into a prompt, denoted as \textit{CP} and feed it to the LLMs. Instead of training a dedicated reward model for the ASR task, we simplify the process by using the sum of log-probabilities. Specifically, for each hypothesis, we compute the total log-probability of the hypothesis based on the LLMs' conversational format as follows:
\texttt{<\textbar{}user\textbar{}>Generate a message optimized for \{CP\} <\textbar{}end\textbar{}><\textbar{}assistant\textbar{}> \{hypothesis\} }.

This prompt encourages the LLMs to bias its generation toward the target domain. For the hypothesis $a_{i,n}$ for audio $X_i$, we calculate the sum of log-probabilities from both the LLM and the ASR model. The reward function is defined as: 

\begin{equation}\label{attention}
  \begin{aligned}
r_{i,n}=P_{llm}(a_{i,n}| CP) + \lambda P_{asr}(a_{i,n}| X_i)
\end{aligned}
\end{equation}

where $\lambda$ is a tunable parameter that controls the contribution of the acoustic model's score in the final reward.

\subsection{Model Finetuning}
We explored several representative RL algorithms used for the ASR model finetuning.
\subsubsection{Rejection sampling fine-tuning}
We follow the rejection sampling fine-tuning (RAFT) in the literature \cite{touvron2023llama, xiong2025minimalist}. In the generating rewards step, we retains only the hypothesis with the highest reward and we only aggregate those positive samples into $D_{pos}$ to fine-tune the model.

The objective function is to maximize the log-likelihood over the dataset $D_{pos}$.

\begin{equation}\label{attention}
  \begin{aligned}
L_{RAFT}=\sum_{a \in {D_p}} log\pi(a|X)
\end{aligned}
\end{equation}

\subsubsection{DPO}
The DPO algorithm \cite{rafailov2023direct} relies on pair-wise preference data. Specifically, two distinct hypotheses are independently sampled from the policy model. These responses are then ranked based on human or AI annotator preferences, resulting in a preferred response $a^+$ and a less preferred one $a^-$. In our framework, we select the hypothesis with the hightest reward as the preferred response and the one with the lowest reward as the less preferred one. This process is repeated multiple times to construct the preference dataset $D = \{X, a^+, a^-\}_n$.

DPO then optimizes the following objective function:
\begin{equation}\label{attention}
  \begin{aligned}
L_{DPO}=-log\sigma(\beta log \frac{\pi(a^+|X)}{\pi_{ref}(a^+|X)} - \beta log \frac{\pi(a^-|X)}{\pi_{ref}(a^-|X)})
\end{aligned}
\end{equation}

where $\pi_{ref}$ denotes the reference model, which in our case is the pretrained ASR model. The original DPO algorithm \cite{rafailov2023direct} is trained on offline, off-policy data. In contrast, our framework always uses the latest parameters to sample hypotheses and trains on on-policy data.

\subsubsection{GRPO}
We also explored to use the GRPO algorithm \cite{shao2024deepseekmath} for model finetuning. GRPO is a simplified variant of PPO \cite{schulman2017proximal} that eliminates the value model and computes rewards via rule-based or model-based methods. It calculates the advantage based on the relative rewards of the outputs within each group. In our setup, we compute the following advantage for the i-th response.
\begin{equation}\label{attention}
  \begin{aligned}
A(a_i, X) = \frac{r_i - mean(r_1, ..., r_n)}{std(r_1, ..., r_n)}
\end{aligned}
\end{equation}

The advantage is essentially the normalized reward. Specifically, $mean(r_1, r_2, ..., r_n)$ is the average of rewards which is often referred to as the baseline in the RL literature, while $std(r_1, r_2, ..., r_n)$ denotes the standard deviation of rewards. This normalization can serve to reduce the variance of the stochastic gradient. And then we use the objection function below for optimization:

\begin{align*}
L_{\text{GRPO}} &= \frac{1}{|D|} \sum_{a \in D} \frac{1}{|a|} \sum_{i=1}^{|a|}
\big[ \min(s_t, \\
&\quad \text{clip}(s_t, 1 - \epsilon, 1 + \epsilon)) A(a_i, X) - \beta \mathcal{D}_KL[\pi||\pi_{ref}] \big]
\end{align*}

where $s_t=\frac{\pi(a|X)}{\pi_{ref}(a|X)}$ is the the importance sampling ratio. $\beta$ is the coefficient of the KL penalty, $\mathcal{D}_KL[\pi||\pi_{ref}]$ is the per-token KL penalty.

\begin{table*}[th]
    \centering
    \begin{tabular}{l cc|cc|cc|cc}
   \toprule
\multirow{2}{*}{\textbf{Model}} & 
\multicolumn{2}{c}{\textbf{Food ordering}} & 
\multicolumn{2}{c}{\textbf{Voice command}} & 
\multicolumn{2}{c}{\textbf{Readability}} &
\multicolumn{2}{c}{\textbf{Average}} \\
\cmidrule(lr){2-3} \cmidrule(lr){4-5} \cmidrule(lr){6-7} \cmidrule(lr){8-9}
 & \textbf{WER} & \textbf{EWER} 
 & \textbf{WER} & \textbf{EWER} 
 & \textbf{WER} & \textbf{EWER} 
 & \textbf{WER} & \textbf{EWER} \\
\midrule
Baseline & 9.67 & 17.79  & 23.77 & 17.02  & 11.50 & 52.14 & 14.98 & 28.98  \\
Self-training & 8.76 & 14.42 & 22.71 & 12.74 &  9.29 & 39.29 & 13.59 & 22.15  \\
\midrule
RAFT & 8.18	& 13.80 & 21.47 & 6.31  & 8.13 & 32.14 & 12.59& \textbf{17.42} \\
DPO & 8.34	& 14.16 & 20.64 & 7.14 & 8.48 & 32.14 & \textbf{12.49} & 17.81 \\
GRPO & 8.41	& 14.16	 & 20.34 & 8.21  & 8.71 & 32.14 & \textbf{12.49} & 18.17 \\
\bottomrule
\\
\end{tabular}
    \caption{Unsupervised domain adaptation benchmark}
    \label{adaptationset}
\end{table*}


\section{Experiments}
This section describes the experimental methodology used to evaluate the effectiveness of the proposed methods for unsupervised adaptation tasks.

\subsection{ASR Baseline}
The baseline ASR model is the standard AED model consists of an encoder network with 24 conformer layers, a decoder network with 6 transformer layers. The encoder and decoder have the embedding dimension 1024, and number of head is 16. The output vocabulary size is 6000 sentence pieces. The ASR model has around 500M parameters. And the training data contains 60k hours of internal English data.

\subsection{RL Setup}
For hypothesis candidate sampling, we use the ASR model described above along with sampling-based beam search to generate responses. The beam width is set to 8 and patience score\cite{kasai2022call} is set to 3 to encourage diversity among hypotheses. For each utterance, the system generates 24 hypotheses, and we subsequently sample 8 responses from them. All models are trained for 3 epochs using the AdamW optimizer on 8 A100 GPUs if not specified otherwise. For the reward function, we use Phi-4-mini \cite{abdin2024phi} as the LLM to compute the LLM-based scores. The ASR-based scores are derived from the same ASR model used for sampling.

\subsection{Unsupervised Domain Adaption}

 To evaluate our method in unsupervised domain adaptation setting, we test our method on 3 different internal customization datasets, where only unlabeled audio data is provided during training. The first dataset focuses on the Starbucks food ordering scenario,  with 300 adaptation utterances. The test set includes approximately 500 utterances containing domain-specific entities. The contextual prompt used for LLM scoring is: "Generate a message for ordering at Starbucks." The second dataset involves voice assistant commands, where users issue instructions such as controlling devices, checking the weather, or setting alarms. The test set consists of 800 utterances and around 5k words, including many domain-relevant entities. The contextual prompt used is: "Generate a voice command message." The third dataset is designed for display-format number recognition, where the goal is for the ASR model to produce more readable outputs for numerical content. The training set contains 300 utterances, and the test set has 200 utterances, all of which include numbers in display-friendly formats such as dates, addresses, or phone numbers, for example, "Call me at (425)-123-4567". The contextual prompt used is: "Generate a message that includes numbers or digits in a display-friendly format."


The experiment results are presented in Table\ref{adaptationset}, where we report word error rate (WER), and entity word error rate (EWER). The “Baseline” row shows the performance of the original ASR model before any adaptation. The “Self-training” row corresponds to a standard self-training approach, where the hypothesis with the highest ASR score is used for fine-tuning. As expected, this method yields improved performance over the baseline. 
The second part of the table presents results from our proposed method, which shares the same data collection and reward generation pipeline but employs different reinforcement learning (RL) algorithms. All three RL algorithms perform similarly across the three evaluation sets. This contrasts with previous findings in the literature, where the state-of-the-art methods GRPO have been shown to outperform others \cite{xiong2025minimalist}. We attribute this discrepancy to the relatively simple nature of the ASR task, which likely requires less complex reasoning, leading to similar behavior across different RL strategies. We also noticed that our method demonstrates relatively smaller improvements on on the food ordering dataset comparing with the other two. We believe this is due to the presence of rare entities in the food domain that were never sampled during RL training, preventing the model from learning the correct transcriptions. Our best-performing system achieves an average relative reduction of 15\% in WER and 50\% in EWER improvement compared to the baseline. Compared to the self-training approach, it yields an 8\% relative reduction in WER and 21\% in EWER improvement on average.These results highlight the effectiveness of LLM-based feedback for unsupervised adaptation, showing not only improved transcription accuracy but also enhanced readability and usability of the output in real-world applications.

\subsection{Readability Customization}
We also conducted experiments on readability customization, similar to the number/digit setting described earlier. However, in this case, the focus is on improving the overall readability of display-formatted ASR output, including accurate punctuation, casing, and other formatting aspects. For training, we used 100 hours of randomly selected data from the SPGISpeech\cite{o2021spgispeech} training set, and trained the model for one epoch. Evaluation was performed on the SPGISpeech test set. The results are presented in Table\ref{spgi}. Compared to the baseline ASR model, our customized system achieved a 7\% relative improvement in WER and a 5.5\% relative reduction in token error rate (TER). Here a token is defined as a unit separated by spaces in a sentence and thus TER counts punctuation errors and is also case sensitive. Qualitative analysis further confirmed that the adapted ASR outputs better conformed to expected display conventions, such as proper casing, punctuation, and the formatting of alphanumeric strings.

\begin{table}[ht]
    \centering
    \begin{tabular}{l cc}
    \hline
        Model        & WER & TER \\
    \hline
        Baseline      & 2.73 & 7.88  \\
        RAFT          & 2.53 & 7.45 \\
        DPO           & 2.59 & 7.46 \\
        GRPO          & 2.53 & 7.46 \\
    \hline \\
    \end{tabular}
    \caption{WER/TER on the SPGISpeech set}
    \vspace{-0.0cm}
    \label{spgi}
\end{table}

\subsection{Contextual information}
To evaluate the impact of contextual information within our reinforcement learning framework, we conducted an ablation study on the food ordering dataset comparing models trained with and without contextual input. When the prompt given to the LLM was simplified from “Generate a message for ordering at Starbucks.” to “Generate a message.”, the results in Table\ref{ci} showed that incorporating context reduced EWER by 5\% and WER by 6\%. This demonstrates that contextual information enables the model to better capture key domain-specific content. In contrast, removing contextual input from the reward scoring process led to degraded preference alignment, highlighting the crucial role of context in accurate reward modeling with the LLMs.

\begin{table}[th]
    \centering
    \begin{tabular}{l cc}
    \hline
        Model        & WER & EWER\\
     \hline
      Baseline      &  9.67 & 17.79  \\
      RAFT & 8.18 & 13.80  \\
      RAFT w/o contextual info & 8.77 & 14.42  \\
     \hline \\
    \end{tabular}
    \caption{Comparison of performance with and without contextual information}
    \vspace{-0.0cm}
    \label{ci}
\end{table}

\subsection{Comparison with other approaches}
We compared our RL based domain adaptation method against existing pipelines that leverage LLMs, including LLM error correction (re-write)\cite{ma2023can, gu2024denoising, ma2025asr} and LLM re-ranking \cite{hu2023massively, li2023prompting, chen2023large, ogawa2024applying} on the food ordering dataset. For a fair comparison, both approaches used the same Phi4-mini \cite{abdin2024phi} model as the backbone LLM. In Table\ref{pp}, our method outperforms LLM-based post-processing in terms of WER, although it does not surpass it on EWER. Notably, while LLM post-processing can improve certain metrics, it often incurs substantially higher inference costs. In contrast, our approach effectively utilizes LLM feedback as a reward signal to directly fine-tune the ASR model, resulting in a simpler and more efficient system tailored to the target domain. Moreover, our approach is complementary and can be combined with those post-processing techniques to further enhance performance.

\begin{table}[th]
    \centering
    \begin{tabular}{l cc}
    \hline
          Model        & WER & EWER  \\
     \hline
      Baseline      &  9.67 & 17.79   \\
      RAFT & 8.18 & 13.80   \\
    \hline
    LLM-rescore &  9.59 & 13.80 \\
    LLM-rewrite &   8.47& 13.50 \\
     \hline
     \\
    \end{tabular}
    \caption{Comparison with LLMs post-processing approaches}
    \vspace{-0.0cm}
    \label{pp}
\end{table}

\section{Conclusion}
In this work, we present a novel RL framework for unsupervised domain adaptation of ASR systems, leveraging feedback from LLMs as a reward signal. By eliminating the need for labeled data and separately trained reward models, our approach harnesses the broad knowledge and contextual understanding of LLMs to score ASR hypotheses, guiding model fine-tuning via RL algorithms. This method is especially effective at improving recognition of rare named entities and adapting to domain mismatches. Our experiments demonstrate that the proposed approach achieves substantial improvements—up to a 21\% relative gain in entity recognition compared to traditional self-training baselines. Future work includes training a dedicated reward model for ASR task to replace LLMs probability scores and extending this method to multi-modality LLMs.
\bibliographystyle{IEEEtran}
\bibliography{refs}

\end{document}